\documentclass[runningheads]{llncs}
\usepackage{graphicx}
\usepackage{comment}
\usepackage{amsmath,amssymb} 
\usepackage{color}

\usepackage{bm}
\usepackage{xspace}
\usepackage{booktabs}
\usepackage{tabularx}
\usepackage{enumitem}
\usepackage[table,dvipsnames]{xcolor}

\makeatletter
\DeclareRobustCommand\onedot{\futurelet\@let@token\@onedot}
\def\@onedot{\ifx\@let@token.\else.\null\fi\xspace}
\def\eg{\emph{e.g}\onedot} 
\def\ie{\emph{i.e}\onedot} 
 
\def\etc{\emph{etc}\onedot} 
 
\def\etal{\emph{et al}\onedot}

\makeatother

\makeatletter
\newcommand{\printfnsymbol}[1]{\textsuperscript{\@fnsymbol{#1}}}
\makeatother

\newcolumntype{Y}{>{\centering\arraybackslash}X}

\def\hdmap{\mathcal{M}}

\def\actors{\mathcal{A}}
\def\bbox{\boldsymbol{b}}
\def\state{\boldsymbol{s}}

\def\tor4d{ATG4D}
\def\NoNoise{NoNoise}
\def\GaussianNoise{GaussianNoise}
\def\MultimodalNoise{MultimodalNoise}
\def\LearnedNoise{ActorNoise}
\def\ContextNoise{ContextNoise}

\begin{document}
\pagestyle{headings}
\mainmatter
\def\ECCVSubNumber{5368}  

\title{Testing the Safety of Self-driving Vehicles by Simulating Perception and Prediction}
\titlerunning{Perception and Prediction Simulation}
\authorrunning{K. Wong, Q. Zhang, M. Liang, B. Yang, R. Liao, A. Sadat, and R. Urtasun}
\author{
    Kelvin Wong\inst{1, 2}\thanks{Indicates equal contribution. Work done during Qiang's internship at Uber ATG.}
    \and Qiang Zhang\inst{1, 3}\printfnsymbol{1}
    \and Ming Liang\inst{1}
    \and Bin Yang\inst{1, 2}
    \and Renjie Liao\inst{1, 2}
    \and Abbas Sadat\inst{1}
    \and Raquel Urtasun\inst{1, 2}
}
\institute{
    Uber Advanced Technologies Group, Toronto, Canada
    \and University of Toronto, Toronto, Canada
    \and Shanghai Jiao Tong University, Shanghai, China \\
    \email{\{kelvin.wong, ming.liang, byang10, rjliao, asadat, urtasun\}@uber.com} \\
    \email{zhangqiang2016@sjtu.edu.cn}
}

\maketitle


\begin{abstract}

We present a novel method for testing the safety of self-driving vehicles in simulation.
We propose an alternative to sensor simulation, as sensor simulation is expensive and has large
domain gaps.
Instead, we directly simulate the outputs of the self-driving vehicle's
perception and prediction system, enabling realistic motion planning testing.
Specifically, we use paired data in the form of ground truth labels and real
perception and prediction outputs to train a model that predicts what the online system will produce.
Importantly, the inputs to our system consists of high definition maps, bounding
boxes, and trajectories, which can be easily sketched by a test engineer in a matter
of minutes.
This makes our approach a much more scalable solution.
Quantitative results on two large-scale datasets demonstrate that
we can realistically test motion planning using our simulations.

\end{abstract}

\section{Introduction}

Self-driving vehicles (SDVs) have the potential to
become a safer, cheaper, and more scalable form of transportation.
But while great progress has been achieved in the last few decades,
there still remain many open challenges that impede the deployment of these vehicles at scale.
One such challenge concerns how to test the safety of these
vehicles and, in particular, their motion planners~\cite{fan2018,sadat2019}.
Most large-scale self-driving programs in industry use simulation
for this purpose,
especially in the case of testing safety-critical scenarios,
which can be costly---even \emph{unethical}---to perform in the real world.
To this end, test engineers first create a large
bank of test scenarios, each comprised of a high definition (HD) map
and a set of actors represented by bounding boxes and trajectories.
These mocked objects are then given as input to the motion planner.
Finally, metrics computed on the simulation results
are used to assess progress.

However, in order to provide realistic testing, the mocked objects need to
reflect the noise of real perception and prediction\footnote{We use the terms \emph{prediction} and \emph{motion forecasting} interchangeably.}
systems~\cite{luo2018,casas2018,liang2020,zhang2020,wang2020,li2020}.
Unfortunately, existing approaches typically assume
perfect perception or use simple heuristics to generate noise~\cite{gu2015}.
As a result, they yield unrealistic assessments of the
motion planner's safety.
For example, under this testing regime, we will never see the SDV slamming
its brakes due to a false positive detection.

\begin{figure}[!t]
\includegraphics[width=\textwidth]{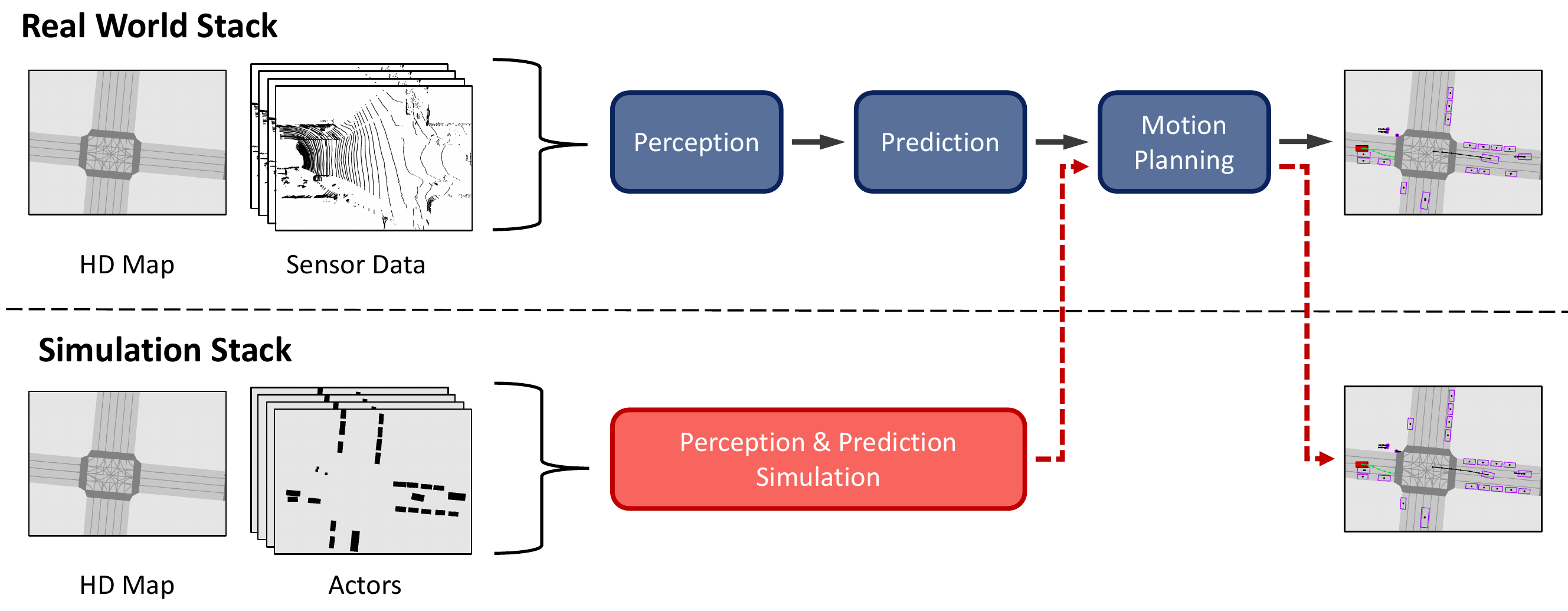}
\caption{\textbf{Perception and prediction simulation.}
Our goal is to simulate the outputs of the SDV's perception and
prediction system in order to realistically test its motion planner.
For each timestep, our system ingests
an HD map and a set of actors (bounding boxes and trajectories) and
produces noisy outputs similar to those from the real system.
To test the motion planner, we mock real outputs with our simulated ones.}
\label{introduction/figure:concept-diagram}
\end{figure}

An alternative approach is to use sensor simulation to test the SDV's
full autonomy stack, end-to-end.
Sensor simulation is a popular area of research, particularly in the case
of images~\cite{ros2016,gaidon2016,alhaija2018,wrennige2018,kar2019,li2019,yang2020}.
However, most existing sensor simulators are costly and difficult to
scale since they are based on virtual worlds created by teams of artists; \eg,
TORCS~\cite{wymann2015}, CARLA~\cite{dosovitskiy2017}, AirSim~\cite{shah2017}.
Rendering these virtual worlds also results in observations that have very
different statistics from real sensor data.
As a result, there are large domain gaps between these virtual worlds
and our physical one.
Recently, LiDARSim \cite{siva2020} leveraged real-world data to produce
realistic LiDAR simulations at scale, narrowing the fidelity gap significantly.
However, current autonomy stacks use a host of different sensors,
including LiDAR~\cite{zhou2018,yang2018,lang2019},
radar~\cite{chadwick2019,yang2020b},
cameras~\cite{chen2016,wang2019,liang2020}, and ultrasonics,
and thus all of these sensors must be simulated consistently for
this approach to be useful in testing the full autonomy stack.
These challenges make sensor simulation a very exciting area of
research, but also one that is potentially far from deployment in real-world systems that
must meet requirements developed by safety, systems engineering, and testing teams.

In this paper, we propose to simulate the SDV's
perception and prediction system instead; see Fig.~\ref{introduction/figure:concept-diagram}.
To this end, we provide a comprehensive study of a variety
of noise models with increasing levels of sophistication.
Our best model is a convolutional neural network that, given a simple
representation of the scene, produces realistic perception and
prediction simulations.
Importantly, this input representation can be sketched by a test engineer in a
matter of minutes, making our approach cheap and easy to scale.
We validate our model on two self-driving datasets and
show that our simulations closely match the outputs of a real
perception and prediction system.
We also demonstrate that they can be used to
realistically test motion planning.
We hope to inspire work in this important field so that
one day we can certify the safety of SDVs and deploy them at scale.


\section{Related Work}

\subsubsection{Sensor simulation:}
The use of sensor simulation in self-driving dates back to at least
the seminal work of Pomerleau~\cite{pomerleau1989} who used both simulated and
real road images to train a neural network to drive.
Since then, researchers and engineers have developed increasingly realistic
sensor simulators for self-driving across various modalities.
For example, \cite{ros2016,gaidon2016,alhaija2018,wrennige2018,kar2019} use
photo-realistic rendering techniques to synthesize images to train
neural networks and \cite{li2019,yang2020} leverage real sensor data to
generate novel views.
Likewise, \cite{blensor2011,yue2018,fang2018,dosovitskiy2017} use
physics-based ray-casting to simulate LiDAR while \cite{siva2020}
enhances its realism with learning.
And in radar, \cite{gubelli2013} propose a ray-tracing based simulator
and \cite{wheeler2017} use a fully-learned approach.
However, despite much progress in recent years, there remain sizeable domain
gaps between simulated sensor data and real ones~\cite{siva2020}.
Moreover, developing a realistic sensor simulator requires significant effort
from domain experts~\cite{kar2019}, which limits the scalability of doing so
across an entire sensor suite.
In this paper, we sidestep these challenges by instead simulating a much
simpler scene representation: the SDV's perception and
prediction outputs.

\subsubsection{Virtual environments:}
Training and testing robots in the phyiscal world can be a slow, costly,
and even dangerous affair;
virtual environments are often used to circumvent these difficulties.
For example, in machine learning and robotics, popular benchmarks include computer
games~\cite{bellemare2013,johnson2016,tessler2017,kempka2016,beattie2016},
indoor environments~\cite{kolve2017,savva2017,gibson2018,gibson2019},
robotics simulators~\cite{coumans2019,todorov2012,koenig2004}, and self-driving
simulators~\cite{wymann2015,chen2015,dosovitskiy2017,shah2017}.
These virtual worlds have motivated a wealth of research in
fields ranging from embodied vision to self-driving.
However, they also require significant effort to construct,
and this has unfortunately limited the diversity of their content.
For example, CARLA~\cite{dosovitskiy2017} originally had
just two artist-generated towns consisting of 4.3km of drivable roads.
In this paper, we use a lightweight scene representation that
simplifies the task of generating new scenarios.

\subsubsection{Knowledge distillation:}
Knowledge distillation was first popularized by Hinton \etal~\cite{hinton2015}
as a way to compress neural networks by training one network with the (soft)
outputs of another.
Since then, researchers have found successful applications of distillation
in subfields across machine learning~\cite{gupta2015,geras2015,papernot2015,rusu2016,guo2018}.
In this paper, we also train our simulation model using outputs from an SDV's
perception and prediction system.
In this sense, our work is closely related with distillation.
However, unlike prior work in distillation, we assume no direct knowledge
of the target perception and prediction system;
\ie, we treat these modules as black boxes.
Moreover, the inputs to our simulation model differ from the inputs to the target system.
This setting is more suitable for self-driving, where perception and prediction
systems can be arbitrarily complex pipelines.

\section{Perception and Prediction Simulation}
\label{section:method}
Our goal is to develop a framework for testing the SDV's motion planner
as it will behave in the real world.
One approach is to use sensor simulation to test the SDV's full autonomy
stack, end-to-end.
However, this can be a complex and costly endeavor that requires constructing
realistic virtual worlds and developing high-fidelity sensor simulators.
Moreover, there remains a large domain gap between the
sensor data produced by existing simulators and our physical world.

In this work, we study an alternative approach.
We observe that the autonomy stack of today's SDVs employ a cascade of
interpretable modules: perception, prediction, and motion planning.
Therefore, rather than simulate the raw sensor data, we
simulate the SDV's intermediate perception and prediction outputs instead,
thus leveraging the compositionally of its autonomy stack to bypass
the challenges of sensor simulation.
Testing the SDV's motion planner can then proceed by simply mocking
real perception and prediction outputs with our simulated ones.
We call this task \emph{perception and prediction simulation}.

Our approach is predicated on the hypothesis that there exists systemic
noise in modern perception and prediction systems that we could simulate.
Indeed, our experiments show that this is the case in practice.
Therefore, we study a variety of noise models with increasing
levels of sophistication.
Our best model is a convolutional neural network that, given a simple
representation of the scene, learns to produce realistic perception and
prediction simulations.
This enables us to realistically test motion planning in simulation.
See Fig.~\ref{introduction/figure:concept-diagram} for an overview.

In this section, we first formulate the task of perception
and prediction simulation and define some useful notation.
Next, we describe a number of noise models in order of increasing sophistication
and highlight several key modeling choices that informs the design
of our best model.
Finally, we describe our best model for this task and
discuss how to train it in an end-to-end fashion.

\subsection{Problem Formulation}
\label{section:method/problem-formulation}

Given a sensor reading at timestep $ t $, the SDV's perception and
prediction system ingests an HD map and sensor data
and produces a class label $ \hat{c}_i $, a bird's eye view (BEV) bounding box $ \hat{\bbox}_i $,
and a set of future states $ \hat{\state}_i = \{\hat{\state}_{i, t + \delta}\}_{\delta = 1}^{H} $
for each actor $ i $ that it detects in the scene, where $ H $ is the prediction horizon.
Each state $ \hat{\state}_{i, t + \delta} \in \mathbb{R}^3 $ consists of the
actor's 2D BEV position and orientation at some timestep $ t + \delta $ in the
future.\footnote{Actors' future orientations are approximated from
their predicted waypoints using finite differences,
and their bounding box sizes remain constant over time.}
Note that this is the typical output parameterization for an SDV's perception
and prediction system~\cite{luo2018,casas2018,liang2020,zhang2020,wang2020,li2020},
as it is lightweight, interpretable, and easily ingested by existing motion planners.

For each timestep in a test scenario, our goal is to simulate the
outputs of the SDV's perception and prediction system without using sensor
data---neither real nor simulated.
Instead, we use a much simpler representation of the world such that
we can: (i) bypass the complexity of developing realistic virtual worlds
and sensor simulators;
and (ii) simplify the task of constructing new test scenarios.

Our scenario representation consists of an HD map
$ \hdmap $, a set of actors $ \actors $, and additional meta-data for motion
planning, such as the SDV's starting state and desired route.
The HD map $ \hdmap $ contains semantic information about the
static scene, including lane boundaries and drivable surfaces.
Each actor $ a_i \in \actors $ is represented by a class label $ c_i $,
a bounding box $ \bbox_i $, and a set of states $ \state_i = \{\state_{i, t}\}_{t = 0}^{T} $,
where $ T $ is the scenario duration.
Note that $ \actors $ is a \emph{perfect} perception and prediction
of the world, not the (noisy) outputs of a real online system.

This simple representation can be easily sketched by
a test engineer in a matter of seconds or minutes, depending on the
complexity and duration of the scenario.
The test engineer can start from scratch or from existing logs collected in real
traffic or in structured tests at a test track by adding or removing actors,
varying their speeds, changing the underlying map, \etc.

\subsection{Perturbation Models for Perception and Prediction Simulation}
\label{section:method/perturbation-models}

\begin{figure}[!t]
\includegraphics[width=\textwidth]{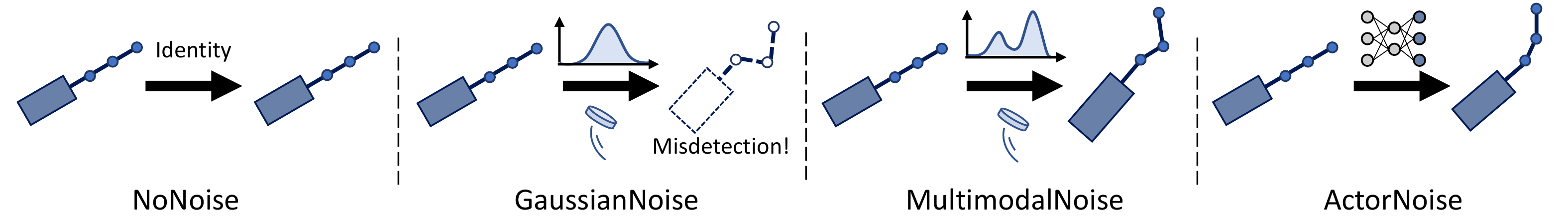}
\caption{\textbf{Perturbation models for perception and prediction simulation.}
{\NoNoise} assumes perfect perception and prediction.
{\GaussianNoise} and {\MultimodalNoise} use marginal noise distributions
to perturb each actor's shape, position, and whether it is misdetected.
{\LearnedNoise} accounts for inter-actor variability by predicting perturbations
conditioned on each actor's bounding box and positions over time.}
\label{formulation/figure:perturbation-architecture}
\end{figure}

One family of perception and prediction simulation methods builds on the
idea of perturbing the actors $ \actors $ of the input test scenario with
noise approximating that found in real systems.
In this section, we describe a number of such methods in order of increasing
sophistication;
see Fig.~\ref{formulation/figure:perturbation-architecture}.
Along the way, we highlight several key modeling considerations that
will motivate the design of our best model.

\subsubsection{\NoNoise:}
For each timestep $ t $ of the test scenario, we can readily simulate perfect
perception and prediction by outputting the class label $ c_i $, the bounding
box $ \bbox_i $, and the future states $ \{\state_{i, t + \delta}\}_{\delta = 1}^{H} $
for each actor $ a_i \in \actors $.
Indeed, most existing methods to test motion planning similarly use
perfect perception~\cite{gu2015}.
This approach gives an important signal as an upper bound on the motion
planner's performance in the real world.
However, it is also unrealistic as it yields an overly
optimistic evaluation of the motion planner's safety.
For example, this approach cannot simulate false negative detections;
thus, the motion planner will never be tested for its ability to exercise
caution in areas of high occlusion.

\subsubsection{\GaussianNoise:}
Due to its assumption of perfect perception and prediction,
the previous method does not account for the noise present in real
perception and prediction systems.
As such, it suffers a sim-to-real domain gap.
In domain randomization, researchers have successfully used random noise to
bridge this gap during training~\cite{peng2017,tobin2017,pouyanfar2019,mehta2019,openai2019}.
This next approach investigates whether random noise can be similarly used to
bridge the sim-to-real gap \emph{during testing.}
Specifically, we model the noise present in real perception and prediction
systems with a marginal distribution $ p_\mathrm{noise} $ over all actors.
For each timestep $ t $ in the test scenario,
we perturb each actor's bounding box $ \bbox_i $ and future states
$ \{\state_{i, t + \delta}\}_{\delta = 1}^{H} $ with noise drawn from $ p_\mathrm{noise} $.
In our experiments, we use noise drawn from a Gaussian distribution $ \mathcal{N}(0, 0.1) $
to perturb each component in $ \bbox_i = (x, y, \log w, \log h, \sin \theta, \cos \theta) $,
where $ (x, y) $ is the box's center, $ (w, h) $ is its width and height,
and $ \theta $ is its orientation.
We similarly perturb each state in $ \{\state_{i, t + \delta}\}_{\delta = 1}^{H} $.
To simulate misdetections, we randomly drop boxes with probability equal to
the observed rate of false negative detections in our data.\footnote{
True positive, false positive, and false negative detections are determined
by IoU following the detection AP metric. In our experiments, we use a 0.5 IoU
threshold for cars and vehicles and 0.3 IoU for pedestrians and bicyclists.}

\subsubsection{\MultimodalNoise:}
Simple noise distributions such as the one
used in {\GaussianNoise} do not adequately reflect the complexity of the
noise in perception and prediction systems.
For example, prediction noise is highly multi-modal since
vehicles can go straight or turn at intersections.
Therefore, in this next approach, we instead use a Gaussian Mixture Model,
which we fit to the empirical distribution of noise in our data via
expectation-maximization~\cite{bishop2006}.
As before, we simulate misdetections by dropping boxes with probability
equal to the observed rate of false negative detections in our data.

\subsubsection{\LearnedNoise:}
In {\MultimodalNoise}, we use a marginal noise distribution over all actors
to model the noise present in perception and prediction systems.
This, however, does not account for inter-actor variability.
For example, prediction systems are usually more accurate for
stationary vehicles than for ones with irregular motion.
In our next approach, we relax this assumption by conditioning the noise for each
actor on its bounding box $ \bbox_i $ and past, present, and future states $ \state_i $.
We implement {\LearnedNoise} as a multi-layer perceptron that
learns to predict perturbations to each component of an actor's
bounding box $ \bbox_i $ and future states
$ \{\state_{i, t + \delta}\}_{\delta = 1}^{H} $.
We also predict each actor's probability of misdetection.
To train {\LearnedNoise}, we use a combination of a binary cross entropy loss for
misdetection classification and a smooth $ \ell_1 $ loss for box and waypoint
regression.

\subsection{A Contextual Model for Perception and Prediction Simulation}
\label{section:method/model-architecture}

\begin{figure}[!t]
\includegraphics[width=\textwidth]{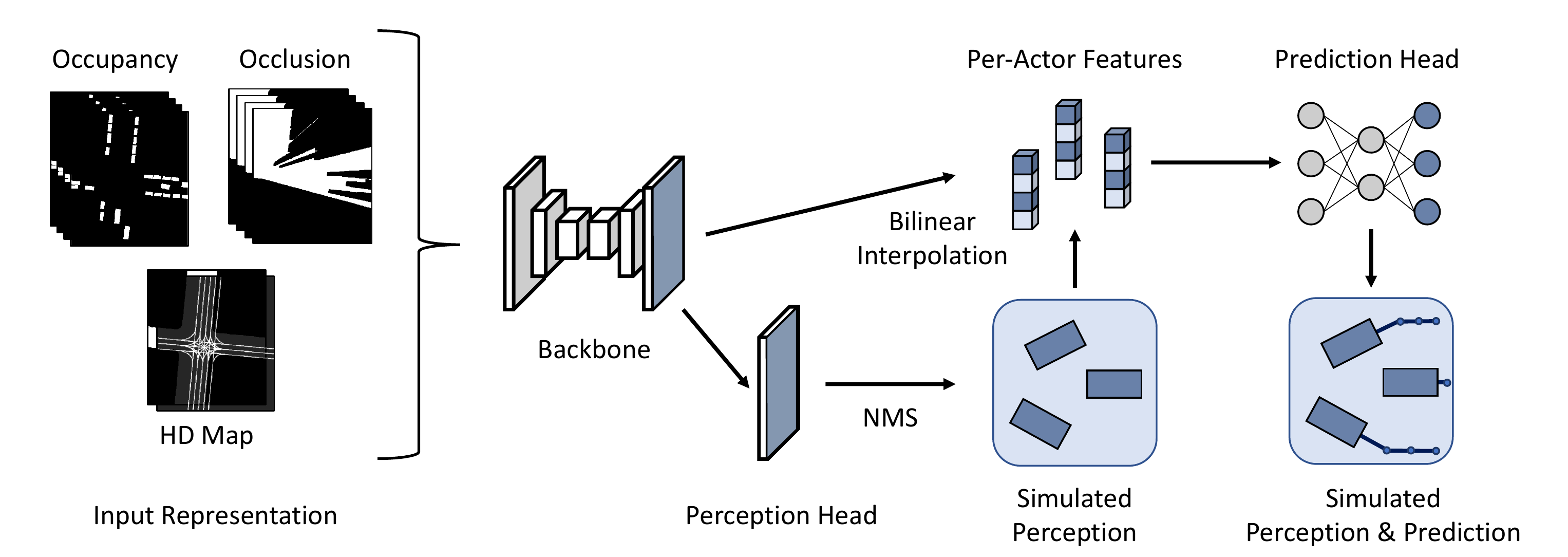}
\caption{\textbf{{\ContextNoise} for perception and prediction simulation.}
Given BEV rasterized images of the scene (drawn from bounding boxes and HD maps),
our model simulates outputs similar to those from the real perception and prediction system.
It consists of: (i) a shared backbone feature extractor;
(ii) a perception head for simulating bounding box outputs; and
(iii) a prediction head for simulating future states outputs.
}
\label{formulation/figure:context-noise-architecture}
\end{figure}

So far, we have discussed several perturbation-based models for
perception and prediction simulation.
However, these methods have two limitations.
First, they cannot simulate false positive misdetections.
More importantly, they do not use contextual
information about the scene, which intuitively should correlate with the success
of a perception and prediction system.
For example, HD maps provide valuable contextual information to determine what
actor behaviors are possible.

To address these limitations, we propose to use a convolutional neural network
that takes as input BEV rasterized images of the scene (drawn from bounding boxes and HD maps)
and learns to simulate dense bounding boxes and future state outputs similar to
those from the real perception and prediction system.
This is the native parameterization of the perception and prediction system
used in our experiments.
Our model architecture is composed of three components:
(i) a shared backbone feature extractor;
(ii) a perception head for simulating bounding box outputs; and
(iii) a prediction head for simulating future states outputs.
We call this model \emph{\ContextNoise}.
See Fig.~\ref{formulation/figure:context-noise-architecture} for an overview.

\subsubsection{Input representation:}
For each timestep $ t $ of the input scenario,
our model takes as input BEV raster images of the scene in ego-centric coordinates.
In particular, for each class of interest, we render the actors of that class
as bounding boxes in a sequence of occupancy masks~\cite{bansal2019,jain2019}
indicating their past, present, and future positions.
Following~\cite{casas2018,yang2018b}, we rasterize the HD map
$ \hdmap $ into multiple binary images.
We represent lane boundaries as polylines and drivable surfaces
as filled polygons.
Occlusion is an important source of systemic errors for perception
and prediction systems.
For example, a heavily occluded pedestrian is more likely to be misdetected.
To model this, we render a temporal sequence of 2D
occlusion masks using a constant-horizon ray-casting algorithm~\cite{gu2015}.
By stacking these binary images along the feature channel, we obtain our
final input representation.

\subsubsection{Backbone network:}
We use the backbone architecture of~\cite{liang2020} as our shared
feature extractor.
Specifically, it is a convolutional neural network that computes a
feature hierarchy at three scales of  input resolution: 1/4, 1/8, and 1/16.
These multi-scale features are then upscaled to 1/4 resolution and fused
using residual connections.
This yields a $ C \times H /4 \times W / 4 $ feature map, where $ C $ is the
number of output channels and $ H $ and $ W $ is the height and width of the
input raster image.
Note that we use this backbone to extract features from BEV raster
images (drawn from bounding boxes and HD maps), \emph{not}
voxelized LiDAR point clouds as it was originally designed for.
We denote the resulting feature map by:
\begin{align}
    \mathcal{F}_\mathrm{bev} = \mathrm{CNN}_\mathrm{bev}\left(\actors, \hdmap\right)
\end{align}

\subsubsection{Perception head:}
Here, our goal is to simulate the bounding box outputs of the real
perception and prediction system.
To this end, we use a lightweight header to predict dense bounding box outputs
for every class.
Our dense output parameterization allows us to naturally handle
false positive and false negative misdetections.
In detail, for each class of interest, we use one convolution layer
with $ 1 \times 1 $ kernels to predict a bounding box $ \tilde{\boldsymbol{b}}_i $
and detection score $ \tilde{\alpha}_i $ at every BEV pixel $ i $ in
$ \mathcal{F}_\mathrm{bev} $.
We parameterize $ \tilde{\boldsymbol{b}}_i $ as
$ (\Delta x, \Delta y, \log w, \log h, \sin \theta, \cos \theta) $,
where $ (\Delta x, \Delta y) $ are the position offsets to the box center,
$ (w, h) $ are its width and height, and $ \theta $ is its orientation~\cite{yang2018}.
We use non-maximum suppression to remove duplicates.
This yields a set of simulated bounding boxes
$ \mathcal{B}_\mathrm{sim} = \{\tilde{\boldsymbol{b}}_i\}_{i = 1}^{N} $.

\subsubsection{Prediction head:}
Our goal now is to simulate a set of future states for
each bounding box $ \tilde{\boldsymbol{b}}_i \in \mathcal{B}_\mathrm{sim} $.
To this end, for each $ \tilde{\boldsymbol{b}}_i \in \mathcal{B}_\mathrm{sim} $,
we first extract a feature vector $ \boldsymbol{f}_i $ by bilinearly interpolating
$ \mathcal{F}_\mathrm{bev} $ around its box center.
We then use a multi-layer perceptron to simulate its future positions:
\begin{align}
    \tilde{\boldsymbol{x}}_{i} = \mathrm{MLP}_{\mathrm{pred}}\left(\boldsymbol{f}_i\right)
\end{align}
where $ \tilde{\boldsymbol{x}}_i \in \mathbb{R}^{H \times 2} $ is a set
of 2D BEV waypoints over the prediction horizon $ H $.
We also simulate its future orientation $ \boldsymbol{\tilde{\theta}}_i $ using finite differences.
Together, $ \{\tilde{\boldsymbol{x}}_i\}_{i = 1}^{N} $ and $ \{\boldsymbol{\tilde{\theta}}_i\}_{i = 1}^{N} $
yield a set of simulated future states
$ \mathcal{S}_\mathrm{sim} = \{\tilde{\boldsymbol{s}_i}\}_{i = 1}^{N} $.
Combining $ \mathcal{S}_\mathrm{sim} $ with
$ \mathcal{B}_\mathrm{sim} $, we have our final perception
and prediction simulation.

\subsubsection{Learning:}
We train our model with a multi-task loss function:
\begin{align}
    \mathcal{L} = \ell_{\mathrm{perc}} + \ell_{\mathrm{pred}}
\end{align}
where $ \ell_\mathrm{perc} $ is the perception loss and
$ \ell_\mathrm{pred} $ is the prediction loss.
Note that these losses are computed between our simulations and the
outputs of the real perception and prediction system.
Thus, we train our model using datasets that provide both real sensor data (to
generate real perception and prediction outputs) and our input scenario
representations (to give as input to our model).\footnote{Our
representation uses bounding boxes and trajectories. Most self-driving
datasets provide this as \emph{ground truth labels} for the standard perception
and prediction task. For perception and prediction simulation, we use
these labels as \emph{inputs} instead.}

Our perception loss is a multi-task detection loss.
For object classification, we use a binary cross-entropy loss with online negative
hard-mining, where positive and negative BEV pixels are determined
according to their distances to an object's center~\cite{yang2018}.
For box regression at positive pixels, we use a smooth $ \ell_1 $ loss
for box orientation and an axis-aligned IoU loss for box location and size.

Our prediction loss is a sum of smooth $ \ell_1 $ losses over future waypoints
for each true positive bounding box, where a simulated box is positive
if its IoU with a box from the real system exceeds a certain threshold.
In our experiments, we use a threshold of $ 0.5 $ for cars and vehicles and
$ 0.3 $ for pedestrians and bicyclists.


\section{Experimental Evaluation}
In this section, we benchmark a variety of noise models for perception
and prediction simulation on two large-scale self-driving datasets (Section~\ref{section:experiments/pnp-results}).
Our best model achieves significantly higher simulation
fidelity than existing approaches that assume perfect perception and prediction.
We also conduct downstream experiments with two motion planners
(Section~\ref{section:experiments/planning-results}).
Our results show that there is a strong correlation between our ability
to realistically simulate perception and prediction and our ability to
realistically test motion planning.

\begin{figure}[!t]
\includegraphics[width=\textwidth]{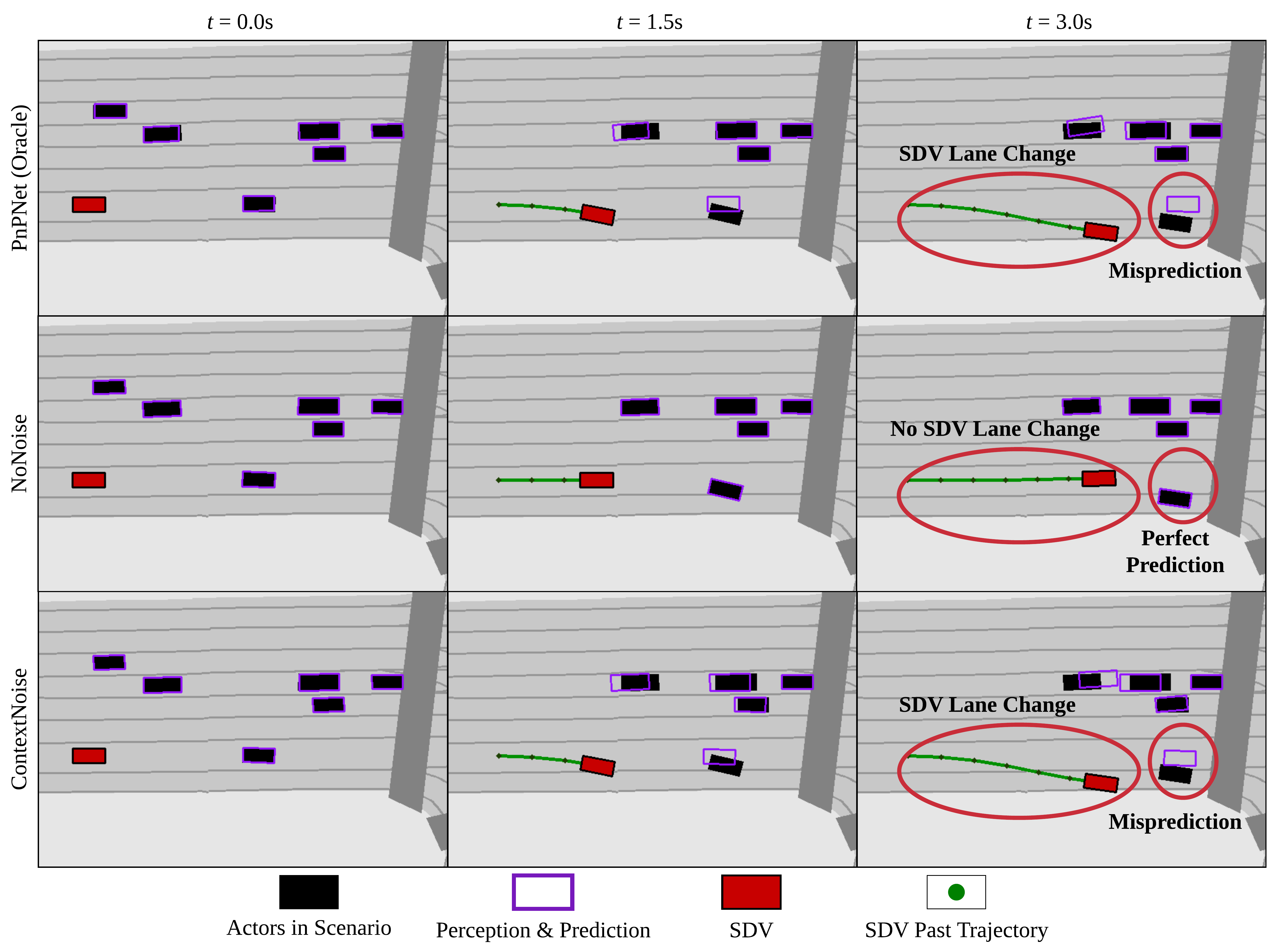}
\caption{\textbf{Simulation results on {\tor4d}.}
We visualize PLT~\cite{sadat2019} motion planning results when given real perception and
prediction (top) versus simulations from {\NoNoise} (middle) and {\ContextNoise} (bottom).
{\ContextNoise} faithfully simulates a misprediction due to multi-modality and
induces a lane-change behavior from the motion planner.}
\label{experiments/figure:qualitative-results}
\end{figure}

\subsection{Datasets}

\subsubsection{nuScenes:}
nuScenes~\cite{nuscenes2019} consists of 1000 traffic scenarios collected in
Boston and Singapore, each containing 20 seconds of video captured by a 32-beam
LiDAR sensor at 20Hz.
In this dataset, keyframes sampled at 2Hz are annotated with object labels
within a 50m radius.
We generate additional labels at unannotated frames by linearly interpolating
labels from adjacent keyframes~\cite{liang2020}.
We use the official training and validation splits
and perform evaluation on the \emph{car} class.
To prevent our simulation model from overfitting to the training split,
we partition the training split into two halves:
one to train the perception and prediction model
and the other our simulation model.
Note that we do not use HD maps in our nuScenes experiments due to
localization issues in some maps.\footnote{As of nuScenes map v1.0.}

\subsubsection{\tor4d:}
{\tor4d}~\cite{yang2018} consists of 6500 challenging traffic scenarios collected
by a fleet of self-driving vehicles in cities across North America.
Each scenario contains 25 seconds of video captured by a
Velodyne HDL-64E at 10Hz, resulting in 250 LiDAR sweeps per video.
Each sweep is annotated with bounding boxes and trajectories for the vehicle,
pedestrian, and bicyclist classes within a 100m radius and comes with localized
HD maps.
We split {\tor4d} into two training splits of 2500 scenarios each, a validation
split of 500, and a test split of 1000.

\subsection{Experiment Setup}

\subsubsection{Autonomy stack:}
We simulate the outputs of PnPNet~\cite{liang2020}---a state-of-the-art
joint perception and prediction model.
PnPNet takes as input an HD map and the past 0.5s of LiDAR sweeps
and outputs BEV bounding boxes and 3.0s of
future waypoints (in 0.5s increments) for each actor that it detects.
Since our focus is on simulating perception and prediction, we use
the variant of PnPNet without tracking.
We configure PnPNet to use a common detection score threshold of 0.
In the {\tor4d} validation split, this corresponds to a
recall rate of 94\% for vehicles, 78\% for pedestrians, and 62\% for bicyclists.

To gauge the usefulness of using our simulations to test motion planning,
we conduct downstream experiments with two motion planners.
Our first motion planner is adaptive cruise control (ACC), which implements a
car-following algorithm.
Our second motion planner is PLT~\cite{sadat2019}---a jointly learnable
behavior and trajectory planner.
PLT is pretrained on the ManualDrive dataset~\cite{sadat2019}, which
consists of 12,000 logs in which the drivers were instructed to drive smoothly.

\subsubsection{Experiment details:}
In nuScenes, we use a $ 100\mathrm{m} \times 100\mathrm{m} $
region of interest centered on the SDV for training and evaluation.
In {\tor4d}, we use one encompassing 70m in front of the
SDV and 40m to its left and right.
Our rasters have a resolution of 0.15625m per pixel, resulting in
$ 640 \times 640 $ input images for nuScenes and $ 448 \times 512 $ for {\tor4d}.
All of our noise models ingest 0.5s of actor states in the past
and 3.0s into the future (in 0.5s increments).
We train {\LearnedNoise} and {\ContextNoise} using the Adam optimizer~\cite{kingma2014}
with a batch size of 32 and an initial learning rate of $ 4\mathrm{e}{-4} $,
which we decay by $ 0.1 $ after every five epochs for a total of 15 epochs.
We re-train PnPNet for our experiments following~\cite{liang2020}.

\subsection{Perception and Prediction Simulation Results}
\label{section:experiments/pnp-results}

In this section, we benchmark a variety of noise models for
perception and prediction simulation.
Our best model, {\ContextNoise}, produces simulations that closely match the
outputs of the real perception and prediction system.

\subsubsection{Metrics:}
We use two families of metrics to evaluate the similarity between our
simulated outputs and those from the real perception and prediction system.
This is possible since our datasets provide both real sensor data and our
input scenario representations.
Our first family of metrics measures the similarity between
simulated bounding boxes and real ones.
To this end, we report detection \emph{average precision} (AP) and \emph{maximum recall}
at various IoU thresholds depending on the class and dataset.
Our second family of metrics measures the similarity between simulated future
states and real ones.
We use \emph{average displacement error} (ADE) over 3.0s and
\emph{final displacement error} (FDE) at 3.0s for this purpose.
These metrics are computed on true positive bounding boxes at 0.5 IoU
for cars and vehicles and 0.3 IoU for pedestrians and bicyclists.
In order to fairly compare models with different maximum recall rates,
we report ADE and FDE for all methods at a common recall point, if it is attained.
All metrics for {\GaussianNoise} and {\MultimodalNoise} are averaged over 25
sample runs.
Note that we use random ordering to compute AP, ADE, and FDE for the methods
that do not produce ranking scores: {\NoNoise}, {\GaussianNoise}, and {\MultimodalNoise}.


\begin{table}[!t]
    \resizebox{\linewidth}{!}{%
    \begin{tabularx}{\textwidth}{ l *{8}{Y}}
    \toprule
                       & \multicolumn{4}{c}{Perception Metrics $ \uparrow $}                & \multicolumn{4}{c}{Prediction Metrics $ \downarrow $}       \\
                       & \multicolumn{2}{c}{AP (\%)}  & \multicolumn{2}{c}{Max Recall (\%)} & \multicolumn{2}{c}{ADE (cm)} & \multicolumn{2}{c}{FDE (cm)} \\
    \midrule
    \textbf{Car}       & 0.5 IoU   & 0.7 IoU          & 0.5 IoU   & 0.7 IoU                 & 50\% R  & 70\% R             & 50\% R     & 70\% R          \\
    \midrule
    ~\GaussianNoise    &  4.9      &  0.9             & 13.0      &  1.7                    & -       & -                  & -          & -               \\
    ~\MultimodalNoise  & 12.8      &  4.9             & 21.1      & 13.1                    & -       & -                  & -          & -               \\
    ~\NoNoise          & 51.5      & 39.0             & 72.0      & 62.7                    & 85      & 84                 & 147        & 146             \\
    ~\LearnedNoise     & 65.7      & 55.0             & 72.1      & 63.5                    & 64      & 66                 & 97         & 100              \\
    ~\ContextNoise     & \bf{72.2} & \bf{59.1}        & \bf{80.3} & \bf{68.9}               & \bf{54} & \bf{61}            & \bf{81}    & \bf{90}         \\
    \bottomrule
    \end{tabularx}%
    }
    \caption{\textbf{Perception and prediction simulation metrics on nuScene validation.}
    \textbf{R} denotes the common recall point at which prediction metrics are computed.}
    \label{table:nuscene-pnp-results}
    \resizebox{\linewidth}{!}{%
    \begin{tabularx}{\textwidth}{ l *{8}{Y}}
    \toprule
                        & \multicolumn{4}{c}{Perception Metrics $ \uparrow $}                  & \multicolumn{4}{c}{Prediction Metrics $ \downarrow $} \\
                        & \multicolumn{2}{c}{AP (\%)}    & \multicolumn{2}{c}{Max Recall (\%)} & \multicolumn{2}{c}{ADE (cm)} & \multicolumn{2}{c}{FDE (cm)}\\
    \midrule
    \textbf{Vehicle}    & 0.5 IoU      & 0.7 IoU         & 0.5 IoU    & 0.7 IoU                & 70\% R    & 90\% R           & 70\% R    & 90\% R \\
    \midrule
    ~\GaussianNoise     & 16.5         &  0.4            & 34.4       &  5.2                   & -         & -                & -         & -       \\
    ~\MultimodalNoise   & 30.7         & 12.1            & 46.8       & 29.4                   & -         & -                & -         & -       \\
    ~\NoNoise           & 71.7         & 56.9            & 93.1       & 82.9                   & 70        & 70               & 127       & 128     \\
    ~\LearnedNoise      & 86.6         & 70.4            & 93.2       & 82.9                   & 65        & 57               & 109       & 93      \\
    ~\ContextNoise      & \bf{91.8}    & \bf{82.3}       & \bf{95.7}  & \bf{87.8}              & \bf{46}   & \bf{51}          & \bf{72}   & \bf{78} \\
    \midrule
    \textbf{Pedestrian} & 0.3 IoU      & 0.5 IoU         & 0.3 IoU    & 0.5 IoU                & 60\% R    & 80\% R           & 60\% R    & 80\% R  \\
    \midrule
    ~\GaussianNoise     & 13.8         &  3.0            & 30.0       & 13.8                   & -         & -                & -         & -       \\
    ~\MultimodalNoise   & 30.3         & 21.7            & 44.2       & 37.4                   & -         & -                & -         & -       \\
    ~\NoNoise           & 57.4         & 52.3            & 84.0       & 80.2                   & 41        & 41               & 70        & 70      \\
    ~\LearnedNoise      & 67.1         & 61.6            & 84.0       & 80.0                   & 36        & 35               & 55        & 54      \\
    ~\ContextNoise      & \bf{75.1}    & \bf{66.6}       & \bf{88.2}  & \bf{80.3}              & \bf{34}   & \bf{34}          & \bf{51}   & \bf{52} \\
    \midrule
    \textbf{Bicyclist}  & 0.3 IoU      & 0.5 IoU         & 0.3 IoU    & 0.5 IoU                & 50\% R    & 70\% R           & 50\% R    & 70\% R  \\
    \midrule
    ~\GaussianNoise     &  4.7         &  0.4            & 17.8       &  5.4                   & -         & -                & -         & -       \\
    ~\MultimodalNoise   &  8.4         &  3.2            & 24.0       & 14.7                   & -         & -                & -         & -       \\
    ~\NoNoise           & 30.6         & 21.6            & 79.7       & 66.8                   & 54        & 55               & 95        & 95      \\
    ~\LearnedNoise      & 60.4         & 44.1            & 79.7       & 67.8                   & 54        & \bf{49}          & 88        & 78      \\
    ~\ContextNoise      & \bf{66.8}    & \bf{52.8}       & \bf{89.8}  & \bf{76.5}              & \bf{52}   & 50               & \bf{80}   & \bf{75} \\
    \bottomrule
    \end{tabularx}%
    }
    \caption{\textbf{Perception and prediction simulation metrics on {\tor4d} test.}}
    \label{table:tor4d-pnp-results}
\end{table}

\subsubsection{Quantitative results:}
Tables~\ref{table:nuscene-pnp-results} and \ref{table:tor4d-pnp-results}
show the results of our experiments on nuScenes and {\tor4d} respectively.
Overall, {\ContextNoise} attains the best performance.
In contrast, simple marginal noise models such as {\GaussianNoise} and
{\MultimodalNoise} perform worse than the method that uses no noise at all.
This attests to the importance of using contextual information for
simulating the noise in real perception and prediction systems.
In addition, we highlight the fact that only {\ContextNoise} improves maximum
recall over {\NoNoise}.
This is at least in part due to its dense output parameterization, which can
naturally model misdetections due to mislocalization, misclassification, \etc.
Finally, we note that {\ContextNoise}'s improvements in
prediction metrics are most evident for the car and vehicle classes;
for rarer classes, such as pedestrians and bicyclists,
{\ContextNoise} and {\LearnedNoise} perform similarly well.

\subsection{Motion Planning Evaluation Results}
\label{section:experiments/planning-results}


\begin{table}[!t]
    \resizebox{\linewidth}{!}{%
    \begin{tabularx}{\textwidth}{ l *{8}{Y}}
    \toprule
                  & \multicolumn{3}{c}{$ \ell_2 $ Distance (cm) $ \downarrow $}  & \multicolumn{2}{c}{Collision Sim. (\%) $ \uparrow $}  & \multicolumn{3}{c}{Driving Diff. (\%) $ \downarrow $} \\
                  & 1.0s     & 2.0s     & 3.0s                        & IoU       & Recall                                    & Beh.       & Jerk      & Acc.        \\ 
    \midrule
    \textbf{PLT} \\
    \midrule
    ~\GaussianNoise    & 2.6      & 8.4      & 15.9                       & 34.5      & 92.7                                      & 0.30       & 0.10      & 1.05      \\ 
    ~\MultimodalNoise  & 2.7      & 9.4      & 18.0                       & 25.2      & \bf{93.6}                                 & 0.33       & 1.22      & 1.25      \\ 
    ~\NoNoise          & 1.4      & 4.8      & 9.5                        & 52.9      & 58.2                                      & 0.18       & 0.44      & \bf{0.03} \\ 
    ~\LearnedNoise     & 1.0      & 3.6      & 7.0                        & 57.6      & 63.6                                      & 0.12       & 0.27      & 0.13      \\ 
    ~\ContextNoise     & \bf{0.8} & \bf{2.9} & \bf{5.6}                   & \bf{65.1} & 74.3                                      & \bf{0.10}  & \bf{0.05} & 0.06      \\ 
    \midrule
    \textbf{ACC} \\
    \midrule
    ~\GaussianNoise   & 6.4      & 32.5     & 79.9                       & 36.5      & \bf{96.7}                                 & -          & 5.14      & \bf{0.03} \\ 
    ~\MultimodalNoise & 5.1      & 26.2     & 64.9                       & 36.5      & \bf{96.7}                                 & -          & 3.84      & 0.11      \\ 
    ~\NoNoise         & 1.9      & 10.0     & 25.2                       & 52.9      & 32.4                                      & -          & 0.20      & 0.17      \\ 
    ~\LearnedNoise    & 1.6      &  8.1     & 20.0                       & 58.6      & 66.3                                      & -          & 0.40      & 0.13      \\ 
    ~\ContextNoise    & \bf{1.4} & \bf{7.2} & \bf{17.6}                  & \bf{61.3} & 74.1                                      & -          & \bf{0.14} & \bf{0.03} \\ 
    \bottomrule
    \end{tabularx}%
    }
        \caption{\textbf{Motion planning evaluation metrics on {\tor4d} test.}}

    \label{table:tor4d-planning-results}
\end{table}

Our ultimate goal is to use perception and prediction simulation to test
motion planning.
Therefore, we conduct downstream experiments in {\tor4d} to quantify the
efficacy of doing so for two motion planners: ACC and PLT.

\subsubsection{Metrics:}
Our goal is to evaluate how similarly a motion planner will behave in
simulation versus the physical world.
To quantify this, we compute the $ \ell_2 $ distance between a motion planner's
trajectory given simulated perception and prediction outputs versus its
trajectory when given real outputs instead.
We report this metric for $ \{1.0, 2.0, 3.0\} $ seconds into the future.
In addition, we also measure their differences in terms of passenger comfort
metrics; \ie, jerk and lateral acceleration.
Finally, we report the proportion of scenarios in which PLT
chooses a different behavior when given simulated outputs instead of real ones.\footnote{
Note that ACC always uses the same driving behavior.}

An especially important metric to evaluate the safety of a motion planner
measures the proportion of scenarios in which the SDV will collide with an obstacle.
To quantify our ability to reliably measure this in simulation, we report the
intersection-over-union of collision scenarios and its recall-based variant:
\begin{align}
    \mathrm{IoU}_{\mathrm{col}} = \frac{|R_+ \cap S_+|}{|R_+ \cap S_+| + |R_+ \cap S_-| + |R_- \cap S_+|}
    & &
    \mathrm{Recall}_\mathrm{col} = \frac{|R_+ \cap S_+|}{|R_+|}
\end{align}
where $ R_+ $ and $ S_+ $ are the sets of scenarios in which the SDV collides
with an obstacle after 3.0s given real and simulated perception and prediction
respectively, and $ R_- $ and $ S_- $ are similarly defined for scenarios with
no collisions.

\subsubsection{Quantitative results:}
Table~\ref{table:tor4d-planning-results} shows our experiment results on {\tor4d}.
They show that by realistically simulating the noise in real perception and
prediction systems, we can induce similar motion planning behaviors in simulation
as in the real world, thus making our simulation tests more realistic.
For example, {\ContextNoise} yields a 41.1\% and 30.2\% relative reduction
in $ \ell_2 $ distance at 3.0s over {\NoNoise} for PLT and ACC respectively.
Importantly, we can also more reliably measure a motion planner's collision
rate using {\ContextNoise} versus {\NoNoise}.
This is an important finding since existing methods to test motion planning
in simulation typically assume perfect perception or use simple heuristics
to generate noise.
Our results show that more sophisticated noise modeling is necessary.


\begin{table}[!t]
\begin{tabularx}{\textwidth}{ c c c c *{8}{Y}}
\toprule
         & \multicolumn{3}{c}{Inputs}           & \multicolumn{3}{c}{AP (\%) $ \uparrow $} & \multicolumn{3}{c}{FDE (cm) $ \downarrow $} & \multicolumn{2}{c}{$ \ell_2 $ @ 3.0s (cm) $ \downarrow $}  \\
Variant  & A          & O          & M          & Veh.      & Ped.      & Bic.             & Veh.     & Ped.    & Bic.                   & PLT      & ACC                                    \\
\midrule
1        & \checkmark &            &            & 85.0      & 64.0      & 59.9             & 87       & 56      & \bf{70}                & 4.7      & 15.0                                    \\
2        & \checkmark & \checkmark &            & 85.5      & 63.8      & 61.7             & 86       & 55      & 72                     & 4.7      & 14.4                                    \\
3        & \checkmark & \checkmark & \checkmark & \bf{86.9} & \bf{68.6} & \bf{64.1}        & \bf{76}  & \bf{52} & \bf{70}                & \bf{4.2} & \bf{14.2}                               \\
\bottomrule
\end{tabularx}
\caption{\textbf{Ablation of ContextNoise input features on {\tor4d} validation.}
We progressively add each input feature described in Section~\ref{section:method/model-architecture}.
\textbf{A} denotes actor occupancy images; \textbf{O} denotes occlusion masks;
and \textbf{M} denotes HD maps.
AP is computed using 0.7 IoU for vehicles and 0.5 IoU for pedestrians and bicyclists.
FDE at 3.0s is computed at 90\% recall for vehicles, 80\% for pedestrians, and 70\% for bicyclists.}
\label{table:input-ablation}
\end{table}

\subsection{Ablation Study}

To understand the usefulness of contextual information for simulation,
we ablate the inputs to {\ContextNoise} by progressively augmenting
it with actor occupancy images, occlusion masks, and HD maps.
From Table~\ref{table:input-ablation}, we see that adding contextual
information consistently improves simulation performance.
These gains also directly translate to more realistic evaluations of motion planning.

\subsection{Qualitative Results}
We also visualize results from the PLT motion planner when given real perception
and prediction versus simulations from {\NoNoise} and {\ContextNoise}.
As shown in Fig.~\ref{experiments/figure:qualitative-results}, {\ContextNoise}
faithfully simulates a misprediction due to multi-modality and induces a
lane-change behavior from the motion planner---the same behavior as if the
motion planner was given real perception and prediction.
In contrast, {\NoNoise} induces an unrealistic keep-lane behavior instead.


\section{Conclusion}

In this paper, we introduced the problem of perception and prediction
simulation in order to realistically test motion planning.
To this end, we have studied a variety of noise models.
Our best model has proven to be a convolutional neural network that,
given a simple representation of the scene, learns to produce realistic
perception and prediction simulations.
Importantly, this representation can be easily sketched by a test engineer
in a matter of minutes.
We have validated our model on two large-scale self-driving datasets and showed
that our simulations closely match the outputs of real perception and
prediction systems.
We have only begun to scratch the surface of this task.
We hope our findings here will inspire advances in this important field so
that one day we can certify the safety of self-driving vehicles
and deploy them at scale.

%
%
\bibliographystyle{splncs04}
\bibliography{egbib}
\end{document}


\pagestyle{headings}
\mainmatter
\def\ECCVSubNumber{5368}  

\title{Supplementary Materials: Testing the Safety of Self-driving Vehicles by Simulating Perception and Prediction} 
\titlerunning{Supplementary Materials: Perception and Prediction Simulation}
\authorrunning{K. Wong, Q. Zhang, M. Liang, B. Yang, R. Liao, A. Sadat, and R. Urtasun}
\author{
    Kelvin Wong\inst{1, 2}\thanks{Indicates equal contribution. Work done during Qiang's internship at Uber ATG.}
    \and Qiang Zhang\inst{1, 3}\printfnsymbol{1}
    \and Ming Liang\inst{1}
    \and Bin Yang\inst{1, 2}
    \and Renjie Liao\inst{1, 2}
    \and Abbas Sadat\inst{1}
    \and Raquel Urtasun\inst{1, 2}
}
\institute{
    Uber Advanced Technologies Group, Toronto, Canada
    \and University of Toronto, Toronto, Canada
    \and Shanghai Jiao Tong University, Shanghai, China \\
    \email{\{kelvin.wong, ming.liang, byang10, rjliao, asadat, urtasun\}@uber.com} \\
    \email{zhangqiang2016@sjtu.edu.cn}
}

\makeatletter
\DeclareRobustCommand\onedot{\futurelet\@let@token\@onedot}
\def\@onedot{\ifx\@let@token.\else.\null\fi\xspace}
\def\eg{\emph{e.g}\onedot} \def\Eg{\emph{E.g}\onedot}
\def\ie{\emph{i.e}\onedot} \def\Ie{\emph{I.e}\onedot}
\def\cf{\emph{c.f}\onedot} \def\Cf{\emph{C.f}\onedot}
\def\etc{\emph{etc}\onedot} \def\vs{\emph{vs}\onedot}
\def\wrt{w.r.t\onedot} \def\dof{d.o.f\onedot}
\def\etal{\emph{et al}\onedot}
\def\resp{\emph{resp}\onedot}
\makeatother

\maketitle

\begin{abstract}
In this document, we provide additional details to supplement the main text.
We first describe additional experiment details (Sec.~\ref{section:additional-experiment-details}).
Then, we provide additional quantitative results that study our approach's
generalization performance (Sec.~\ref{section:structured-test-generalization})
and the effect of the training dataset's size on final performance (Sec.~\ref{section:training-datset-ablation}).
Finally, in Sec.~\ref{section:additional-qualitative-results},
we present a number of qualitative results that demonstrate
the efficacy of using perception and prediction simulation for testing
motion planning.
\end{abstract}

\section{Additional Experiment Details}
\label{section:additional-experiment-details}

\subsection{Model Architectures}

\subsubsection{\MultimodalNoise.}
We implement {\MultimodalNoise} as a Gaussian Mixture Model with $ k = 8 $
components, each with a full covariance matrix.
We use the Scikit-learn implementation~\cite{scikit-learn}.
We also model misdetection noise by fitting a Bernoulli distribution to the
rate of false negative detections in our training split.

\subsubsection{\LearnedNoise.}
{\LearnedNoise} takes as input the actor's bounding box parameters
$ (x, y, w, h, \theta) $, where $ (x, y) $ is the box's center,
$ (w, h) $ are the box's width and height,
and $ \theta $ is the box's heading angle,
as well as the actor's past and future positions centered at $ (x, y) $.
This input feature vector is then processed by a multi-layer perceptron.
In particular, we use a model architecture consisting of:
(i) an initial fully-connected layer with 128-dimensional hidden features,
ReLU activations~\cite{glorot2011}, and group normalization~\cite{wu2018};
(ii) two fully-connected residual blocks~\cite{he2016} with 128-dimensional
hidden features, ReLU activations, and group normalization; and
(iii) a final fully-connected layer to predict perturbations to
the actor's bounding box and future states as well as a misdetection score.

\subsubsection{\ContextNoise.}
As we discussed in the main text, {\ContextNoise} consists of three components:
(i) a shared backbone feature extractor;
(ii) a perception head to simulate bounding box outputs; and
(iii) a prediction head to simulate future states outputs.
We adapt the backbone network architecture described in~\cite{liang2020} to
process raster image inputs and output a 4x downsampled 256-dimensional feature map.
Our perception head is a single 2D convolution layer with $ 1 \times 1 $ kernels
and our prediction head is a multi-layer perceptron adapted from the
architecture used in {\LearnedNoise}.
We use non-maximum suppression thresholds of 0.5 IoU for the cars and vehicles
and 0.3 IoU for the pedestrians and bicyclists.

\section{Additional Quantitative Results}

\subsection{Generalization to Structured Test Scenarios}
\label{section:structured-test-generalization}


\begin{table}[t]
    \resizebox{\linewidth}{!}{%
    \begin{tabularx}{\textwidth}{ l *{8}{Y}}
    \toprule
                        & \multicolumn{4}{c}{Perception Metrics $ \uparrow $}                & \multicolumn{4}{c}{Prediction Metrics $ \downarrow $}       \\
                        & \multicolumn{2}{c}{AP (\%)}  & \multicolumn{2}{c}{Max Recall (\%)} & \multicolumn{2}{c}{ADE (cm)} & \multicolumn{2}{c}{FDE (cm)} \\
    \midrule
    \textbf{Vehicle}    & 0.5 IoU   & 0.7 IoU          & 0.5 IoU   & 0.7 IoU                 & 70\% R  & 90\% R             & 70\% R     & 90\% R          \\
    \midrule
    ~\NoNoise           & 77.2      & 71.9             & 98.0      & \bf{94.5}               & 73      & 73                 & 141        & 141             \\
    ~\ContextNoise      & \bf{94.5} & \bf{88.7}        & \bf{98.4} & 93.4                    & \bf{54} & \bf{54}            & \bf{96}    & \bf{92}         \\
    \midrule
    \textbf{Pedestrian} & 0.3 IoU   & 0.5 IoU          & 0.3 IoU   & 0.5 IoU                 & 60\% R  & 80\% R             & 60\% R     & 80\% R          \\
    \midrule
    ~\NoNoise           & 47.4      & 46.5             & 82.1      & 81.4                    & \bf{36} & \bf{36}            & 64         & 64            \\
    ~\ContextNoise      & \bf{81.9} & \bf{78.8}        & \bf{96.7} & \bf{93.0}               & 37      & 38                 & \bf{60}    & \bf{61}         \\
    \midrule
    \textbf{Bicyclist}  & 0.3 IoU   & 0.5 IoU          & 0.3 IoU   & 0.5 IoU                 & 50\% R  & 70\% R             & 50\% R     & 70\% R          \\
    \midrule
    ~\NoNoise           & 47.8      & 45.2             & 99.5      & 96.7                    & \bf{77} & \bf{77}            & 143        & 143             \\
    ~\ContextNoise      & \bf{89.6} & \bf{87.2}        & \bf{99.6} & \bf{97.8}               & 79      & 83                 & \bf{132}   & \bf{140}         \\
    \bottomrule
    \end{tabularx}%
    }
    \\
    \\
    \\
    \resizebox{\linewidth}{!}{%
    \begin{tabularx}{\textwidth}{ l *{8}{Y}}
    \toprule
                       & \multicolumn{3}{c}{$ \ell_2 $ Distance (cm) $ \downarrow $}  & \multicolumn{2}{c}{Collision Sim. (\%) $ \uparrow $}  & \multicolumn{3}{c}{Driving Diff. (\%) $ \downarrow $} \\
                       & 1.0s     & 2.0s     & 3.0s                        & IoU      & Recall                                    & Beh.       & Jerk      & Acc.        \\ 
    \midrule
    \textbf{PLT} \\
    \midrule
    ~\NoNoise          & 1.2      & 3.5      & 5.6                        & 70.3      & 72.8                                      & 0.32       & 0.69      & 1.71         \\ 
    ~\ContextNoise     & \bf{0.7} & \bf{1.7} & \bf{2.6}                   & \bf{85.2} & \bf{90.4}                                 & \bf{0.08}  & \bf{0.06} & \bf{0.01}    \\ 
    \midrule
    \textbf{ACC} \\
    \midrule
    ~\NoNoise         & 1.7      & 8.0       & 19.8                       & 62.2      & 62.6                                      & -          & 0.26      & 0.46      \\ 
    ~\ContextNoise    & \bf{1.4} & \bf{6.3}  & \bf{15.1}                  & \bf{76.8} & \bf{77.8}                                 & -          & \bf{0.17} & \bf{0.17} \\ 
    \bottomrule
    \end{tabularx}%
    }
    \caption{\textbf{Generalization to structured test scenarios.}
    We evaluate {\NoNoise} and {\ContextNoise} (trained on {\tor4d}) on 500
    logs of structured tests collected at a test track.
    \textbf{R} denotes the common recall point at which prediction metrics are computed.}
    \label{table:tvt-results}
\end{table}

In order to study our approach's generalization performance to novel
interesting-to-test scenarios, we evaluate {\NoNoise} and {\ContextNoise}
(trained on {\tor4d}) on 500 logs of structured tests collected at a test track.
These logs contain rare and safety-critical scenarios that are commonly used
to evaluate self-driving vehicles.
Note that this dataset is selectively labeled; that is, we only annotate
actors that might interact with the SDV.
As such, our noise models are given only these actors as input.
To ensure a fair comparison, we compute metrics comparing our simulations
against real perception and prediction outputs that are \emph{near}\footnote{
A detected actor is \emph{near} an annotated actor if the intersection-over-union
between their bounding boxes exceed 0.1\%.} an annotated actor.

Our results are shown in Table~\ref{table:tvt-results}.
They indicate that {\ContextNoise} generalizes to these novel scenarios
and produces more realistic perception and prediction simulations than {\NoNoise}.
Importantly, they also show that {\ContextNoise} enables more realistic testing
of motion planning in simulation.
This gives us confidence to use perception and prediction simulation to evaluate
motion planning in many variations of these scenarios created by adding or
removing actors, varying their speeds, changing the underlying map, \etc.
It is cost-prohibitive and unsafe to do the same with real-world testing.

\subsection{Effect of Training Dataset Size}
\label{section:training-datset-ablation}


\begin{table}[t]
    \resizebox{\linewidth}{!}{%
    \begin{tabularx}{\textwidth}{ l *{8}{Y}}
    \toprule
                         & \multicolumn{4}{c}{Perception Metrics $ \uparrow $}                & \multicolumn{4}{c}{Prediction Metrics $ \downarrow $}       \\
    \% of Training Split & \multicolumn{2}{c}{AP (\%)}  & \multicolumn{2}{c}{Max Recall (\%)} & \multicolumn{2}{c}{ADE (cm)} & \multicolumn{2}{c}{FDE (cm)} \\
    \midrule
    \textbf{Vehicle}     & 0.5 IoU   & 0.7 IoU          & 0.5 IoU   & 0.7 IoU                 & 70\% R  & 90\% R             & 70\% R     & 90\% R          \\
    \midrule
    ~{\NoNoise} (0\%)    & 61.1      & 54.3             & 93.1      & 87.8                    & 70      & 70                 & 132        & 131             \\
    ~5\%                 & 88.0      & 80.3             & 93.8      & 87.4                    & 74      & 77                 & 123        & 127             \\
    ~10\%                & 89.3      & 82.7             & 94.6      & 88.6                    & 64      & 69                 & 108        & 114             \\
    ~25\%                & 90.6      & 84.6             & 95.1      & 89.5                    & 53      & 57                 & 86         & 91              \\
    ~50\%                & 91.1      & 85.5             & 95.2      & 89.8                    & 46      & 52                 & 73         & 81              \\
    ~100\%               & \bf{91.7} & \bf{86.9}        & \bf{95.4} & \bf{90.7}               & \bf{44} & \bf{49}            & \bf{70}    & \bf{76}         \\
    \midrule
    \textbf{Pedestrian}  & 0.3 IoU   & 0.5 IoU          & 0.3 IoU   & 0.5 IoU                 & 60\% R  & 80\% R             & 60\% R     & 80\% R          \\
    \midrule
    ~{\NoNoise} (0\%)    & 50.8      & 47.4             & 82.8      & 80.0                    & 40      & 40                 & 69         & 69              \\
    ~5\%                 & 69.6      & 59.0             & 84.7      & 75.8                    & 50      & 52                 & 78         & 82              \\
    ~10\%                & 70.2      & 61.1             & 85.2      & 77.6                    & 43      & 45                 & 66         & 69              \\
    ~25\%                & 71.4      & 63.1             & 85.2      & 77.2                    & 36      & 38                 & 55         & 57              \\
    ~50\%                & 73.4      & 65.7             & 86.0      & 78.7                    & 34      & 35                 & \bf{51}    & 53              \\
    ~100\%               & \bf{75.2} & \bf{68.6}        & \bf{86.6} & \bf{80.3}               & \bf{33} & \bf{34}            & \bf{51}    & \bf{52}         \\
    \midrule
    \textbf{Bicyclist}   & 0.3 IoU   & 0.5 IoU          & 0.3 IoU   & 0.5 IoU                 & 50\% R  & 70\% R             & 50\% R     & 70\% R         \\
    \midrule
    ~{\NoNoise} (0\%)    & 33.4      & 29.8             & 83.2      & 78.3                    & \bf{49} & 48                 & 88         & 87             \\
    ~5\%                 & 62.4      & 47.8             & 89.4      & 74.3                    & 126     & 118                & 204        & 190            \\
    ~10\%                & 68.0      & 55.0             & 90.6      & 77.5                    & 90      & 81                 & 146        & 129            \\
    ~25\%                & 70.5      & 60.6             & 91.0      & 81.4                    & 64      & 62                 & 101        & 96             \\
    ~50\%                & 72.5      & 63.7             & 92.2      & 82.7                    & 55      & 51                 & 87         & 78             \\
    ~100\%               & \bf{74.1} & \bf{64.0}        & \bf{92.4} & \bf{82.5}               & \bf{49} & \bf{46}            & \bf{75}    & \bf{70}        \\
    \bottomrule
    \end{tabularx}%
    }
    \\
    \\
    \\
    \resizebox{\linewidth}{!}{%
    \begin{tabularx}{\textwidth}{ l *{8}{Y}}
    \toprule
                         & \multicolumn{3}{c}{$ \ell_2 $ Distance (cm) $ \downarrow $}    & \multicolumn{2}{c}{Collision Sim. (\%) $ \uparrow $}  & \multicolumn{3}{c}{Driving Diff. (\%) $ \downarrow $} \\
    \% of Training Split & 1.0s     & 2.0s     & 3.0s                        & IoU        & Recall                                    & Beh.       & Jerk      & Acc.        \\ 
    \midrule
    \textbf{PLT} \\
    \midrule
    ~{\NoNoise} (0\%)    & 1.2      & 3.9      & 7.5                         & 58.5       & 65.7                                      & 0.16       & 0.29      & 0.41        \\ 
    ~5\%                 & 1.1      & 3.6      & 6.7                         & 64.9       & 77.0                                      & 0.11       & 0.26      & 0.16        \\ 
    ~10\%                & 1.2      & 4.1      & 7.9                         & 67.0       & 83.2                                      & 0.14       & 0.91      & \bf{0.10}   \\ 
    ~25\%                & 0.8      & 2.6      & 4.9                         & 69.2       & \bf{86.0}                                 & 0.10       & \bf{0.07} & 0.12        \\ 
    ~50\%                & \bf{0.7} & 2.4      & 4.6                         & 74.0       & 83.2                                      & 0.09       & 0.34      & 0.23        \\ 
    ~100\%               & \bf{0.7} & \bf{2.2} & \bf{4.2}                    & \bf{74.9}  & 85.4                                      & \bf{0.07}  & 0.22      & 0.20        \\ 
    \midrule
    \textbf{ACC} \\
    \midrule
    ~{\NoNoise} (0\%)    & 1.4      & 7.3      & 18.5                        & 52.3       & 53.3                                      & -          & 0.40      & 0.27        \\ 
    ~5\%                 & 1.7      & 9.1      & 22.9                        & 58.1       & 68.9                                      & -          & 0.93      & 0.15        \\ 
    ~10\%                & 1.5      & 7.9      & 20.0                        & 59.5       & 73.6                                      & -          & 0.36      & \bf{0.04}   \\ 
    ~25\%                & 1.3      & 6.8      & 16.8                        & 63.1       & 74.1                                      & -          & \bf{0.10} & 0.15        \\ 
    ~50\%                & \bf{1.1} & 5.8      & 14.7                        & 63.1       & \bf{88.2}                                 & -          & 0.84      & 0.36        \\ 
    ~100\%               & \bf{1.1} & \bf{5.6} & \bf{14.2}                   & \bf{64.3}  & 82.3                                      & -          & 0.38      & 0.29        \\ 
    \bottomrule
    \end{tabularx}%
    }
    \caption{\textbf{Ablation of training dataset size on {\tor4d} validation.}}
    \label{table:training-size-ablation}
\end{table}

We also study the effects of training dataset sizes on the fidelity of our
simulations.
To this end, we train five {\ContextNoise} models on progressively smaller
subsets of the {\tor4d} training split, starting from 2500 scenarios (100\%) to
125 scenarios (5\%), and we evaluate their performance on the full {\tor4d}
validation split.
We also evaluate {\NoNoise}, which serves as a baseline
for a method that does not require training.

Table~\ref{table:training-size-ablation} shows our results.
Unsurprisingly, {\ContextNoise}'s simulation fidelity is positively correlated
with the size of the training dataset.
It is worth noting, however, that even with much less training data,
{\ContextNoise} still retains good simulation fidelity.
For example, $ \mathrm{IoU}_\mathrm{col} $ decreases by just 0.9\%
(\resp, 1.2\%) in absolute terms for PLT (\resp, ACC) when the training dataset
is halved.
We also observe that the effect of the training dataset's size on validation
performance varies by class---{\ContextNoise} requires much fewer training
scenarios in order to outperform {\NoNoise} on common classes like vehicles
than on rare classes like bicyclists.

\section{Additional Qualitative Results}
\label{section:additional-qualitative-results}

\begin{figure}[t]
  \includegraphics[width=\textwidth]{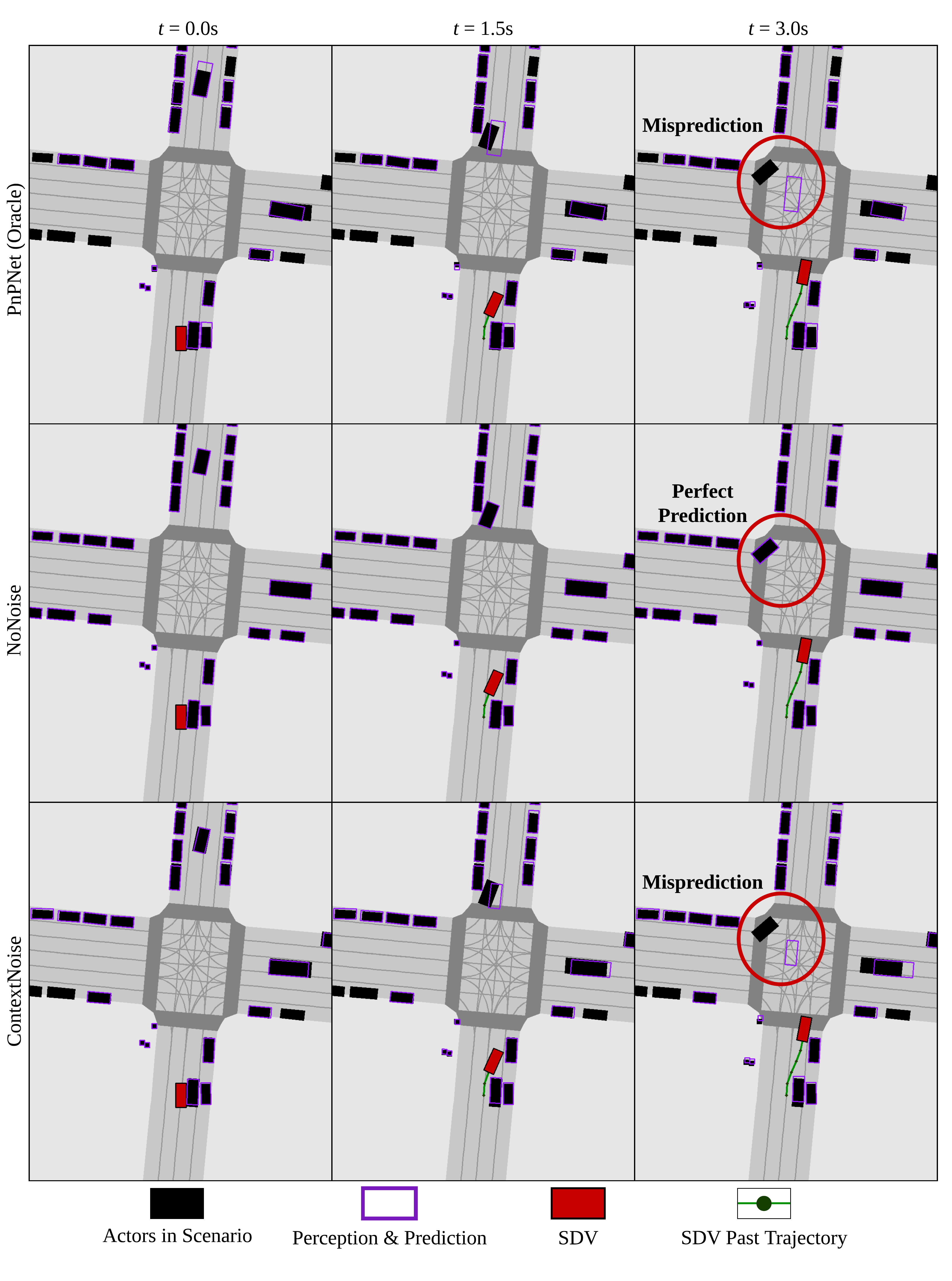}
  \caption{\textbf{Simulating mispredictions.}
  We demonstrate {\ContextNoise}'s ability to simulate mispredictions due to multi-modality.
  Here, both PnPNet and {\ContextNoise} depict the highlighted vehicle as going
  straight when it is in fact turning right.
  {\NoNoise} cannot simulate such mispredictions.}
  \label{figure:cca511d7-65a7-4f46-cd93-e3422828f8e7_183}
\end{figure}

\begin{figure}[t]
  \includegraphics[width=\textwidth]{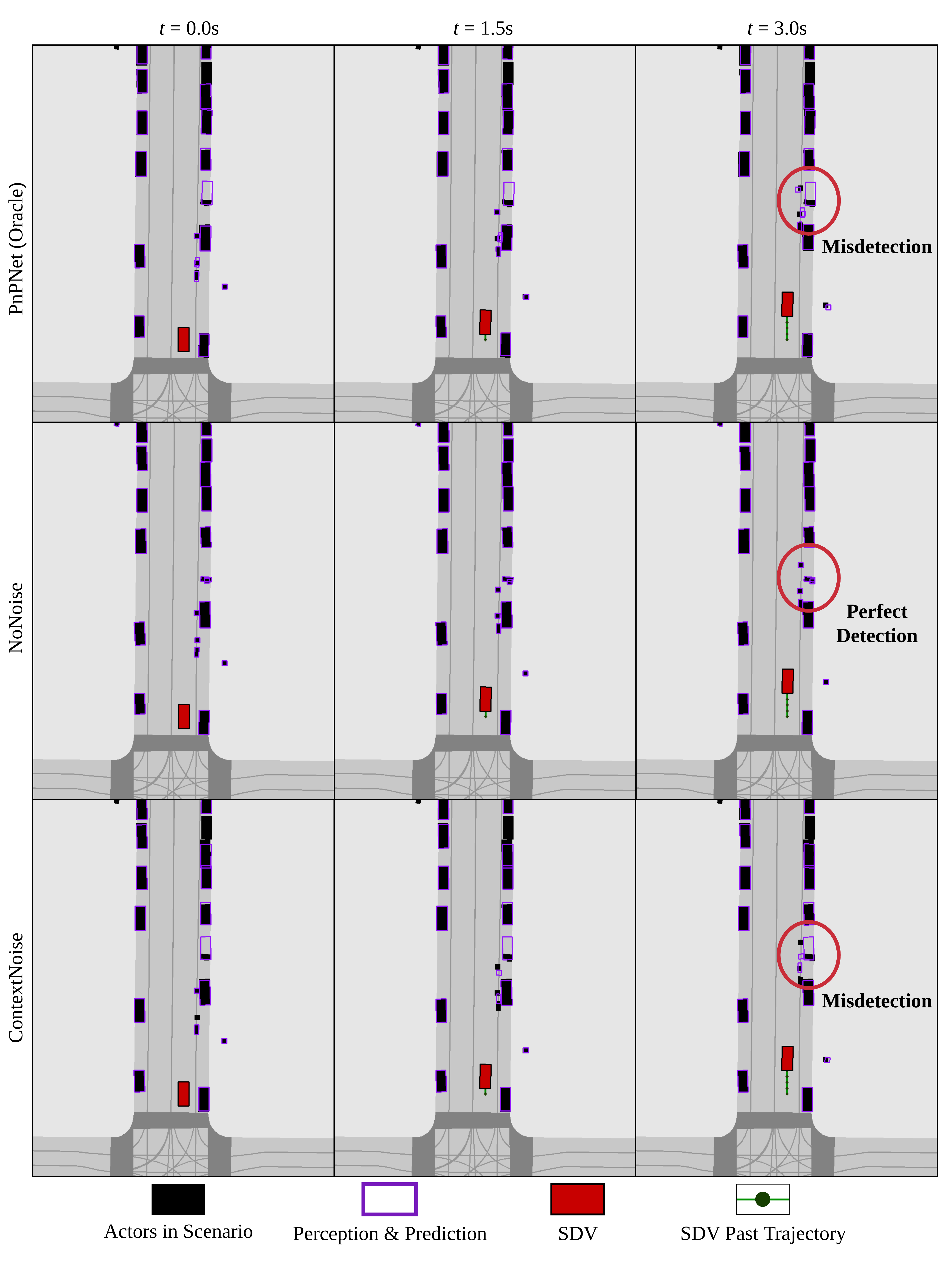}
  \caption{\textbf{Simulating misdetections.}
  We demonstrate {\ContextNoise}'s ability to simulate misdetections due to occlusion.
  In particular, both PnPNet and {\ContextNoise} depict the highlighted
  pedestrians as a parked vehicle.
  {\NoNoise} fails to simulate this misdetection.}
  \label{figure:694bea99-6952-4c85-fcd6-293926bc2bfa_046}
\end{figure}

In Figs.~\ref{figure:cca511d7-65a7-4f46-cd93-e3422828f8e7_183}
to~\ref{figure:0297bbc1-901e-4b8d-e063-d632de449442_144},
we exhibit a number of qualitative results on the {\tor4d} dataset.
Each figure depicts the perception, prediction, and motion planning
outputs for one frame of a scenario, which we unroll over three seconds.
We depict perception and prediction outputs as purple boxes and the SDV as
a red box.
We also depict perfect perception and prediction as black boxes and the HD map
as gray elements.

The top row of each figure shows the outputs from the PLT planner~\cite{sadat2019}
given real perception and prediction outputs from PnPNet~\cite{liang2020}.
These outputs represent the oracle for our task since they are precisely
what our simulations aim to emulate.
Note that we obtain these outputs by passing real sensor data through PnPNet and PLT.
This is possible since {\tor4d} provides both sensor data and our scenario
representation (bounding boxes and trajectories) for every scenario.
We emphasize, however, that our simulation approach does not require sensor data.

The middle and bottom row of each figure similarly depicts simulations obtained
using {\NoNoise} and {\ContextNoise} respectively.
{\NoNoise} represents the prevalent approach of using perfect perception and
prediction to test motion planning in simulation~\cite{gu2015} whereas
{\ContextNoise} is our best simulation model.

\subsubsection{Simulating mispredictions.}
In Fig.~\ref{figure:cca511d7-65a7-4f46-cd93-e3422828f8e7_183}, we show an
example of {\ContextNoise} simulating a misprediction due to multi-modality.
Here, both PnPNet and {\ContextNoise} depict the highlighted vehicle as
going straight, when it is in fact turning right.
Since {\NoNoise} assumes perfect perception and prediction, it falsely
depicts the highlighted vehicle as turning right.

\subsubsection{Simulating misdetections.}
In Fig.~\ref{figure:694bea99-6952-4c85-fcd6-293926bc2bfa_046}, we show
an example where {\ContextNoise} faithfully simulates a misdetection due to
occlusion.
In particular, both PnPNet and {\ContextNoise} depict the highlighted pedestrians
as a parked vehicle.
By contrast, {\NoNoise} fails to simulate this misdetection.

\subsubsection{Detecting collisions in simulation.}
\begin{figure}[t]
  \includegraphics[width=\textwidth]{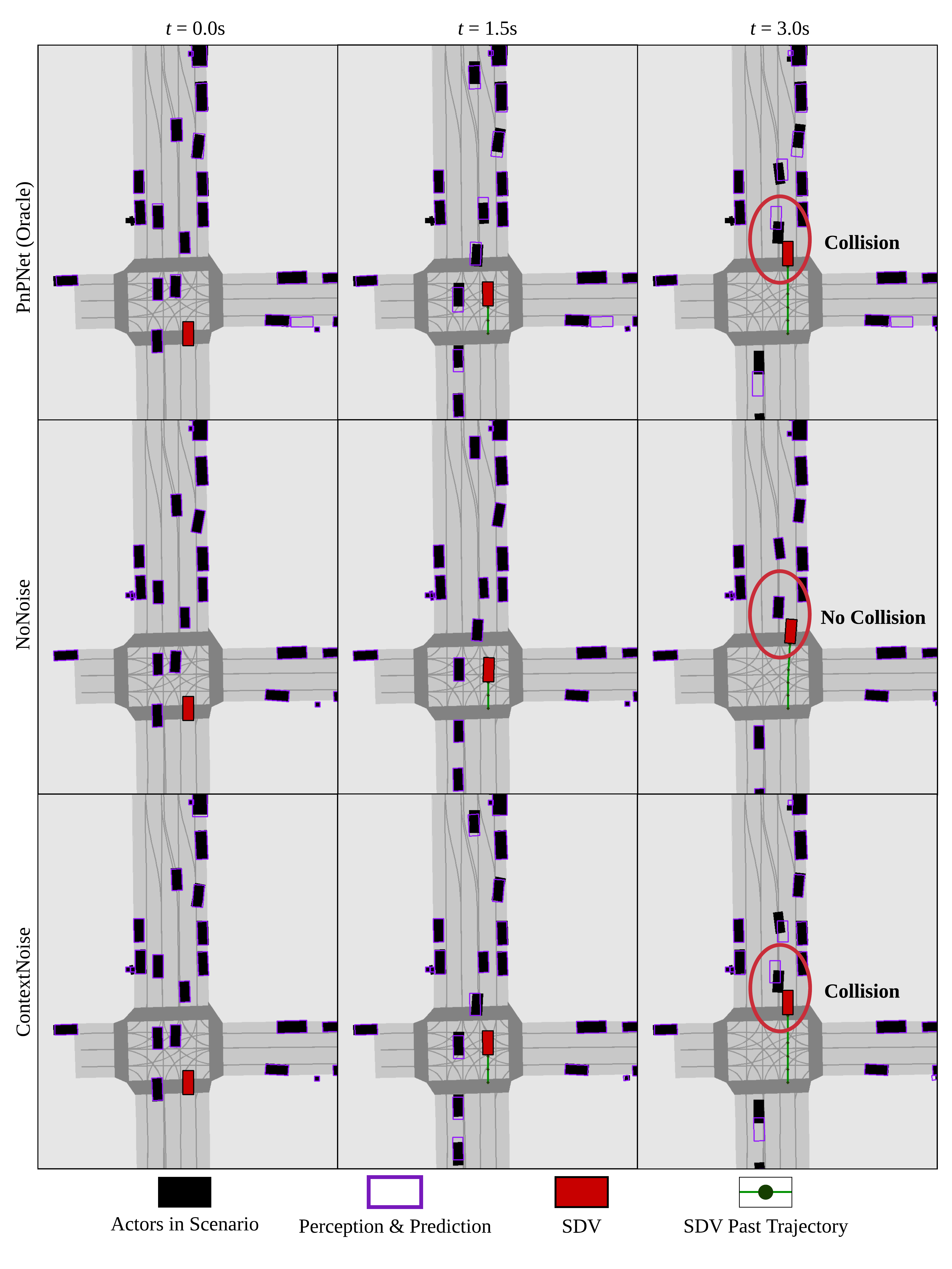}
  \caption{\textbf{Detection collisions in simulation.}
  We show an example in which the PLT motion planner outputs a trajectory that
  results in a collision with a vehicle due to misprediction errors.
  Using simulated outputs from {\ContextNoise}, we can identify this collision
  in simulation.
  In contrast, when given perfect perception and prediction, the motion planner
  safely (but unrealistically) avoids the collision.}
  \label{figure:7670234a-1360-4f29-d821-458c2c4ea4ed_005}
\end{figure}

\begin{figure}[t]
  \includegraphics[width=\textwidth]{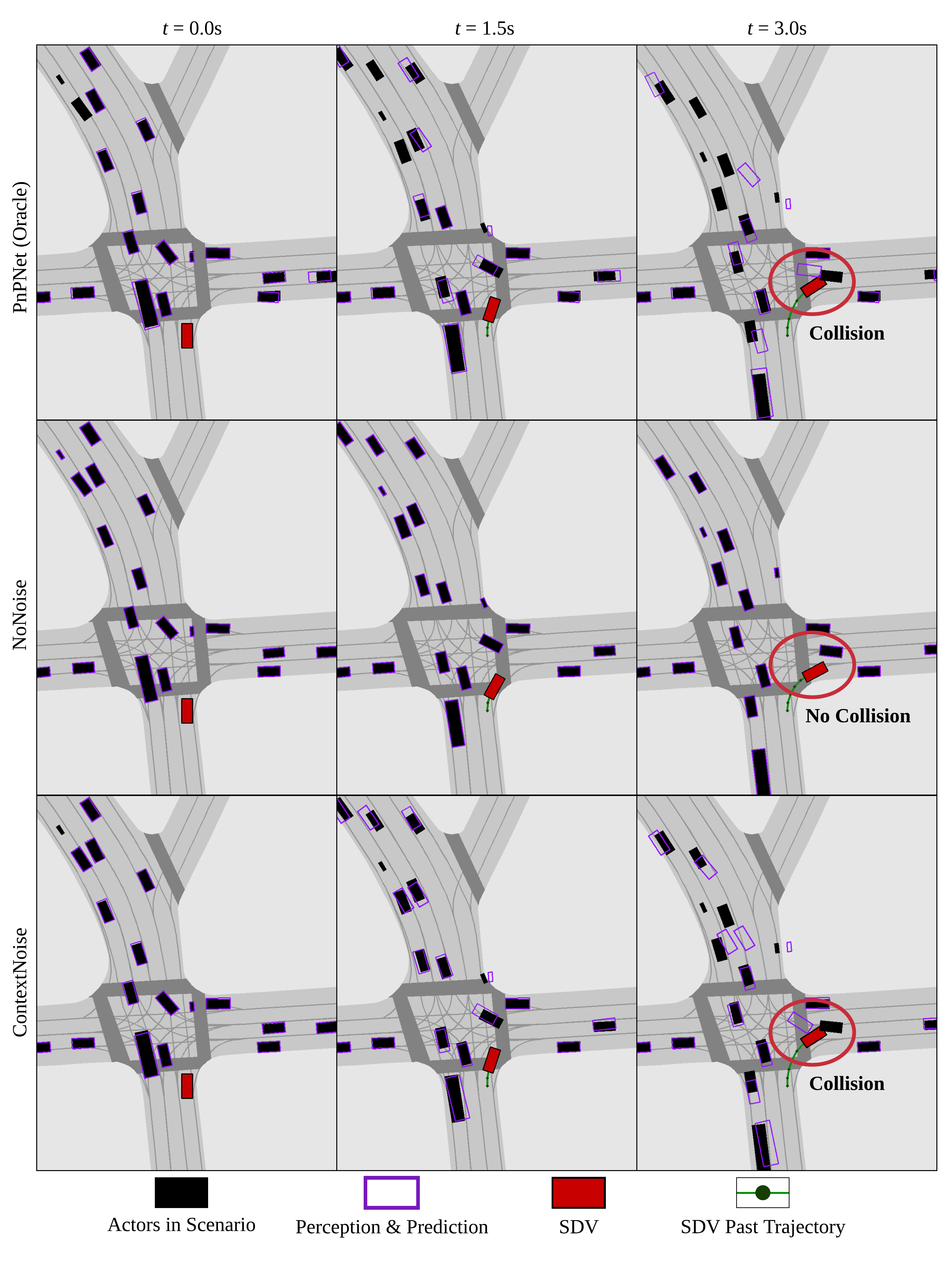}
  \caption{\textbf{Detection collisions in simulation.}
  We show an example in which the PLT motion planner outputs a trajectory that
  results in a collision with a vehicle due to misprediction errors.
  Using simulated outputs from {\ContextNoise}, we can identify this collision
  in simulation.
  In contrast, when given perfect perception and prediction, the motion planner
  safely (but unrealistically) avoids the collision.}
  \label{figure:bd1b637f-0d5e-4160-d203-0f5a84696e41_047}
\end{figure}

\begin{figure}[t]
  \includegraphics[width=\textwidth]{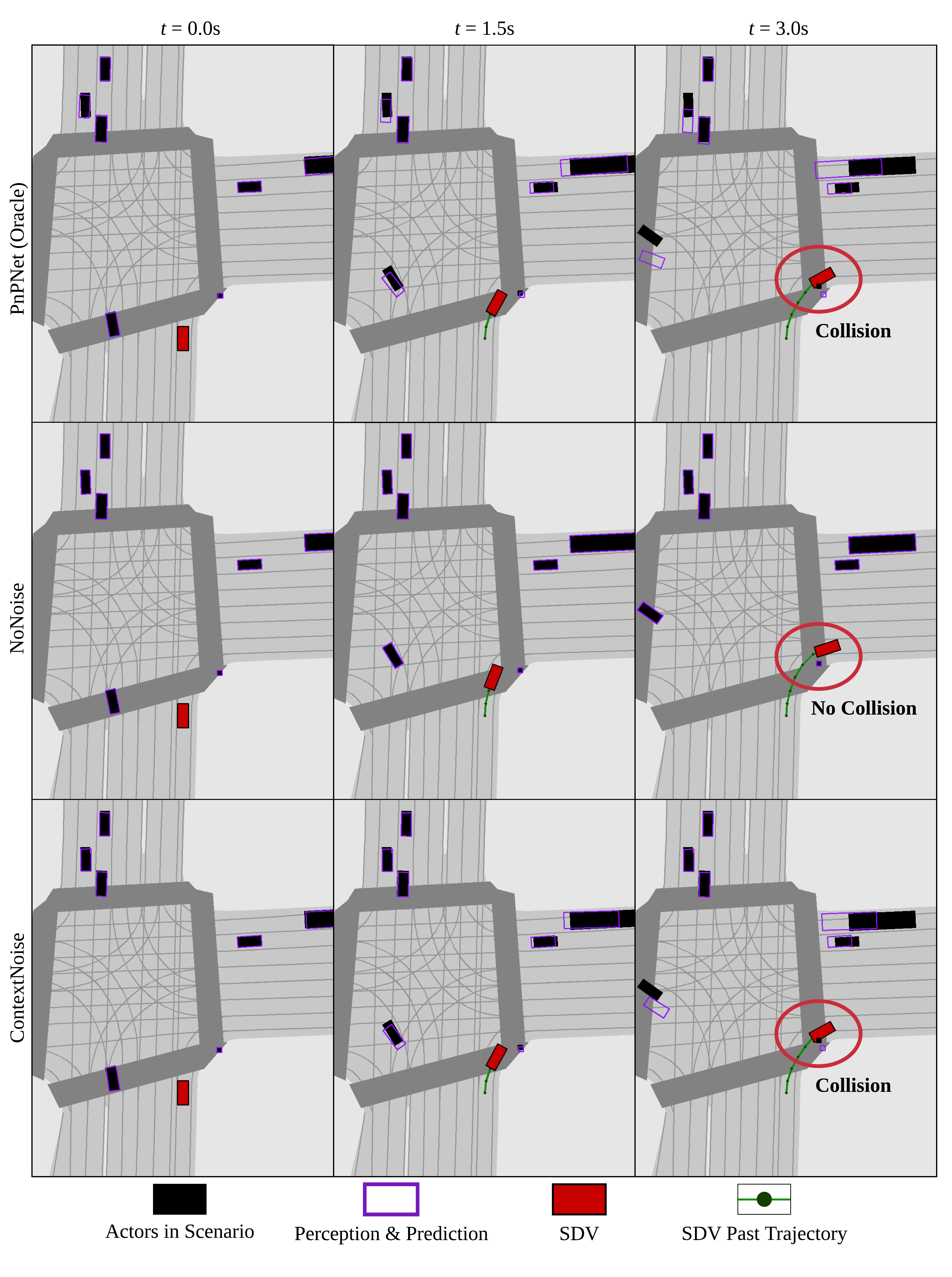}
  \caption{\textbf{Detection collisions in simulation.}
  We show an example in which the PLT motion planner outputs a trajectory that
  results in a collision with a pedestrian due to misprediction errors.
  Using simulated outputs from {\ContextNoise}, we can identify this collision
  in simulation.
  In contrast, when given perfect perception and prediction, the motion planner
  safely (but unrealistically) avoids the collision.}
  \label{figure:0297bbc1-901e-4b8d-e063-d632de449442_144}
\end{figure}

In Figs.~\ref{figure:7670234a-1360-4f29-d821-458c2c4ea4ed_005}
to~\ref{figure:0297bbc1-901e-4b8d-e063-d632de449442_144}, we show
scenarios in which the PLT planner outputs a trajectory that results in
a collision due to misprediction errors.
Figs.~\ref{figure:7670234a-1360-4f29-d821-458c2c4ea4ed_005}
and~\ref{figure:bd1b637f-0d5e-4160-d203-0f5a84696e41_047} show examples
of collisions with vehicles whereas Fig.~\ref{figure:0297bbc1-901e-4b8d-e063-d632de449442_144}
shows an example of a collision with a pedestrian.
By faithfully simulating these misprediction errors with {\ContextNoise}, we
are able to identify these collision scenarios in simulation.
In contrast, when given perfect perception and prediction, the motion planner
safely (but unrealistically) avoids all collisions.
These examples attest to our ability to realistically test motion planning
using simulated outputs from {\ContextNoise}.

%
%
\bibliographystyle{splncs04}
\bibliography{egbib}